\documentclass[review]{elsarticle}

\usepackage{lineno,hyperref}

\usepackage{amssymb}
\newtheorem{definition}{Definition}[section]

\usepackage{amsfonts}
\usepackage{color}
\usepackage{amsmath}
\usepackage{algorithm}
\usepackage{algorithmic}
\usepackage{setspace}
\usepackage{booktabs}
\usepackage{array}

\newtheorem{theorem}{Theorem}

\journal{Journal of \LaTeX\ Templates}

\begin{document}

\begin{frontmatter}

\title{What Objective Does Self-paced Learning Indeed Optimize?}

\author[ad1]{Deyu~Meng\corref{mycorrespondingauthor}}
\cortext[mycorrespondingauthor]{Corresponding author}
\ead{dymeng@mail.xjtu.edu.cn}

\author[ad1]{Qian~Zhao}
\ead{timmy.zhaoqian@gmail.com}


\address[ad1]{School of Mathematics and Statistics and Ministry of Education
Key Lab of Intelligent Networks and Network Security, Xi¡¯an Jiaotong University,
Xian 710049, PR China}

\author[ad2]{Lu Jiang}
\address[ad2]{School of Computer Science, Carnegie Mellon University}
\ead{lujiang@cs.cmu.edu}

\begin{abstract}
Self-paced learning (SPL) is a recently raised methodology designed through simulating the learning principle of humans/animals. A variety of SPL realization schemes have been designed for different computer vision and pattern recognition tasks, and empirically substantiated to be effective in these applications. However, the investigation on its theoretical insight is still a blank. To this issue, this study attempts to provide some new theoretical understanding under the SPL scheme. Specifically, we prove that the solving strategy on SPL accords with a majorization minimization algorithm implemented on a latent objective function. Furthermore, we find that the loss function contained in this latent objective has a similar configuration with non-convex regularized penalty (NSPR) known in statistics and machine learning. Such connection inspires us discovering more intrinsic relationship between SPL regimes and NSPR forms, like SCAD, LOG and EXP. The robustness insight under SPL can then be finely explained. We also analyze the capability of SPL on its easy loss prior embedding property, and provide an insightful interpretation to the effectiveness mechanism under previous SPL variations. Besides, we design a group-partial-order loss prior, which is especially useful to weakly labeled large-scale data processing tasks. Through applying SPL with this loss prior to the FCVID dataset, which is currently one of the biggest manually annotated video dataset, our method achieves state-of-the-art performance beyond previous methods, which further helps supports the proposed theoretical arguments.
\end{abstract}


\end{frontmatter}

\section{Introduction}

Since being raised, \emph{curriculum learning}~(CL)~\cite{CL} and \emph{self-paced learning}~(SPL)~\cite{SPLVM} have been attracting increasing attention in machine learning and pattern recognition communities. The philosophy under this learning paradigm is to simulate the learning principle of humans/animals, which generally starts by learning easier aspects of a learning task, and then gradually takes more complex examples into training \cite{Khan}. Instead of heuristically designing a curriculum by ranking samples based on manually preset easiness measurements as in CL \cite{CLApp1,CLAPP3}, SPL formulates this ad-hoc idea as a concise model through introducing a regularization term into the learning objective. Such amelioration guides a sound SPL regime to automatically optimize an appropriate curriculum by the model itself, which renders it generalize well to various applications and avoids subjective ``easiness" measure setting problem~\cite{SPShift,SPSegmentation,SPVCD,SPTrack}. Very recently, a variety of SPL realization schemes, like self-paced reranking (SPaR)~\cite{SPaR} and self-paced multi-instance learning (SP-MIL)~\cite{SPMIL}, have been proposed and substantiated to be effective to multiple computer vision and multimedia analysis tasks.

Albeit rational in intuition and effective in experience, there exist few research on explaining the underlying mechanism \mbox{inside} SPL. Specifically, even though it is easy to prove that the SPL regime is convergent by adopting an alternative optimization strategy (AOS) on SPL model, it is still unclear where this SPL iteration converges to and why SPL is capable of performing robust learning especially on heavily noisy data. Such in-depth investigations, however, can be considerably necessary for future developments of CL, SPL and their related realizations, and will illuminate whether the SPL methodology would be just an idealistic method occasionally perform on several datasets,  or a rigorous and solid scientific research field worthy to be further explored.

This study aims at understanding the theoretical insight under SPL. Our main results can be summarized as the follows:

First, we prove that the AOS algorithm commonly utilized to solve the SPL problem is identical to a \emph{majorization minimization} (MM)~\cite{MM} algorithm implemented on a latent SPL objective function. In the recent decade, MM has attracted much attention in machine learning and optimization, and many theoretical results on it have been proposed. Such result facilitates an easy analysis on the properties underlying the SPL solving strategy, like convergence and stability, by utilizing the existing knowledge on MM.

Second, we prove that the loss function contained in this latent SPL objective has a close relationship with \emph{non-convex regularized penalty} (NCRP).
Specifically, we discover that multiple current SPL realizations exactly comply with some well known NCRP terms, e.g., the hard and linear SPL regimes are equivalent to the optimizations on losses with the forms of capped-norm penalty (CNP) \cite{CAP1,CAP2,CAP3} and minimax concave plus penalty (MCP) \cite{MCP}, respectively. Such connection inspires us discovering more intrinsic relationship between SPL regimes and known NSPR forms, like smoothly clipped absolute deviation (SCAD)~\cite{SCAD,PR1}, logarithmic penalty (LOG)~\cite{LOG2} and nonconvex exponential penalty (EXP)~\cite{EXP}.

Third, by connecting the SPL optimization with the NCPR loss minimization problems, we provide an easy explanation on why SPL is able to perform robust in the presence of outliers/heavy noises, and accordingly illustrate new insightful understandings on the intrinsic working mechanism under SPL. We also analyze the superiority of SPL on its easy loss prior embedding property beyond conventional learning strategies with pre-fixed loss function. Such property is expected to help a non-convex optimization problem better averting unreasonable local minima and makes SPL more comply with the instructor-student-collaborative-learning mode in human education. Such understanding facilitates an in-depth interpretation for its intrinsic effective mechanism of SPL in previous applications~\cite{SPaR,SPCL,SPLD,SPMF,SPMIL}.

We also propose a group-partial-order loss prior to make SPL better perform in weakly labeled large-scale problems, and implement this SPL regime on the FCVID dataset, which is currently one of the biggest manually annotated video dataset. Our method achieves state-of-the-art performance as compared with previous methods. The results further support the theoretical arguments presented in this study.

The paper is organized as follows. Section 2 introduces related work on this research. Section 3 presents our main theoretical results, and clarify the relationships between AOS and MM algorithms as well as SPL and NCRP problems. Section 4 introduces group-partial-order loss prior and provides the related experimental results on the FCVID dataset. A concluding remark is finally made.

\section{Related Work}

\textbf{Curriculum Learning (CL).}
Inspired by the intrinsic learning principle of humans/animals, Bengio et al. \cite{CL} formalized the fundamental definition of CL. The core idea is to incrementally involve samples into learning, where easy samples are introduced first and more complex ones are then gradually included. These gradually included samples from easy to complex correspond to the curriculums learned in different grown-up stages of humans/animals. This strategy, as supported by empirical evaluation, is helpful in alleviating the local optimum problem in non-convex optimization \cite{CLApp1_,CLA2}.

\textbf{Self-paced Learning (SPL).} Instead of using the heuristic
strategies, Kumar et al. \cite{SPLVM} formulated the key principle of
CL as a concise SPL model. Formally, given a training dataset $\mathcal{D}%
=\{(x_{i},y_{i})\}_{i=1}^{n}$, in which $\mathbf{x}_{i}$ and $y_{i}$ denote the $i^{th}$
observed sample and its label, respectively, $L(y_{i},g(\mathbf{x}%
_{i},\mathbf{w}))$ denotes the loss function which calculates the cost
between the ground truth label $y_{i}$ and the estimated one $g(\mathbf{x}%
_{i},\mathbf{w})$, and $\mathbf{w}$ represents the model parameter in decision function $g$. The SPL model includes a weighted loss term on
\mbox{all} samples and a general self-paced regularizer imposed on sample weights,
expressed as:
\begin{equation}\label{eq:selfpaced_obj}
	\hspace{-4mm}\min_{\mathbf{w},\mathbf{v}\in \lbrack 0,1]^{n}}\!\mathbf{E}(\mathbf{w},\mathbf{v},\lambda)=\sum_{i=1}^{n}\left( v_{i}L(y_{i},f(\mathbf{x}_{i},\mathbf{w}))+f(v_{i},\lambda )\right),
\end{equation}

\noindent
where $\lambda $ is the age parameter for controlling the learning pace, and $f(v,\lambda )$ represents the self-paced regularizer
(SP-regularizer), whose intrinsic conditions have been theoretically abstracted by \cite{SPaR,SPMF}. Through jointly learning the model parameter $\mathbf{w}$ and the latent weight $\mathbf{v}=[v_{1},\cdots ,v_{n}]^{T}$ by AOS with gradually
increasing age parameter, more samples can be automatically included into training from easy to complex in a purely
self-paced way.

Multiple variations of this SPL learning regime, like self-paced reranking~\cite{SPaR}, self-paced multiple instance learning~\cite{SPMIL,SPMIL-PAMI}, self-paced learning with diversity~\cite{SPLD} and self-paced curriculum learning~\cite{SPCL}, have been proposed under the format (\ref{eq:selfpaced_obj}).
The effectiveness of this SPL \mbox{paradigm}, especially its robustness in highly corrupted data, has
been empirically validated in various machine learning and computer vision tasks, such as object detector adaptation \cite{SPShift}, specific-class segmentation learning \cite{SPSegmentation}, visual category discovery \cite{SPVCD}, and long-term tracking \cite{SPTrack}. For example, the SPL paradigm has been a major contributing factor to the leading performance of the CMU team in the challenging TRECVID MED competition organized by the NIST in 2014~\cite{MED14}.

There is few investigation, however, to theoretically explain the intrinsic mechanism under SPL. In this paper, we attempt to enhance the theoretical understanding on this learning paradigm.

\textbf{Non-convex Regularized Penalty (NCRP).}
NCRP has been demonstrated to have attractive properties in sparse estimation (as a penalty term)~\cite{CAP1,MCP,SCAD} and robust learning (as a loss term)~\cite{MCPLoss,CAPLoss} both theoretically and practically, and attracted much attention in machine learning and statistics in recent years. Various NCRP realizations have also been proposed. Typical ones include the capped-norm based penalty (CNP)~\cite{CAP1,CAP2,CAP3}, the minimax concave plus penalty (MCP)~\cite{MCP}, the smoothly clipped absolute deviation penalty (SCAD)~\cite{SCAD},the  logarithmic penalty (LOG)~\cite{LOG2} and the nonconvex exponential penalty (EXP)~\cite{EXP}. The mathematical forms of these NCRP terms in one dimension cases are listed as follows~\cite{NCPList,ZhangNonconvex}:
\begin{equation}\label{eq:NCRP}
\begin{array}{l}
		{\rm CNP:~}p_{\gamma ,\lambda
		}^{CNP}(t)=\gamma \min (|t|,\lambda ),\lambda >0
		\\
		{\rm MCP:~}p_{\gamma ,\lambda }^{MCP}(t)=\left\{
		\begin{array}{c}
			\gamma (|t|-\frac{t^{2}}{2\gamma \lambda }),~{\rm if~}|t|<\gamma \lambda\\
			\frac{\gamma ^{2}\lambda }{2},~{\rm if~}|t|\geq \gamma \lambda
		\end{array}
		\right.
		\\
		{\rm SCAD:~}p_{\gamma ,\lambda }^{SCAD}(t)=\left\{
		\begin{array}{c}
			\lambda |t|,~{\rm if~}|t|\leq \lambda  \\
			\frac{t^{2}-2\gamma \lambda |t|+\lambda ^{2}}{2(1-\gamma)},~{\rm if~}
			\lambda <|t|\leq \gamma \lambda  \\
			\frac{(\gamma +1)\lambda ^{2}}{2},~{\rm if~}|t|\geq \gamma \lambda
		\end{array}
		\right.				\\
{\rm LOG:~}p_{\gamma ,\alpha
		}^{LOG}(t)=\frac{1}{\gamma }\log (1+\alpha |t|)\\
		{\rm EXP:~}p_{\gamma ,\alpha}^{EXP}(t)=\frac{1}{\gamma }(1-\exp (-\alpha |t|)).
	\end{array}
\end{equation}

Albeit possessing elegant statistic properties and empirically verified to be effective in specific applications through finely designed solving strategy,
involving such NCPR terms brings non-convexity to the model. This inclines to result in the issue that the algorithm easily get stuck to an undesired local minima of the problem~\cite{MCPLoss,CAPLoss}.

In this work, we will construct the relationship between the NCRP terms and the SPL regimes, and show that
various helpful loss prior knowledge can be easily embedded into the SPL framework, which is expected to facilitate a NCPR model possibly avoiding unreasonable local minima of the problem and attaining more rational ones better complying with real states.

\textbf{Majorization Minimization Algorithm (MM).}
MM algorithms have wide applications in machine learning and statistical inference~\cite{MM1}. It aims to turn a complicated optimization problem into a tractable one by alternatively iterating the majorization and minimization steps. In particular, considering a minimization problem with objective $F(\mathbf{w})$, given an estimate $\mathbf{w}^k$ at the $k^{th}$ iteration, a typical MM algorithm consists of the following two steps:

\emph{Majorization Step}: Substitute $F(\mathbf{w})$ by a surrogate function $Q(\mathbf{w}|\mathbf{w}^k)$ such that
$$F(\mathbf{w}) \leq Q(\mathbf{w}|\mathbf{w}^k)
$$
with equality holding at $\mathbf{w}=\mathbf{w}^k$.

\emph{Minimization Step}: Obtain the next parameter estimate $\mathbf{w}^{k+1}$ by solving the following minimization problem:
$$
\mathbf{w}^{k+1} = \arg \min_\mathbf{w} Q(\mathbf{w}|\mathbf{w}^k).
$$

\noindent
It is easy to see that when the the minimization of $Q(\mathbf{w}|\mathbf{w}^k)$ is tractable, the MM algorithm can then be very easily implemented, even when the original objective $F(\mathbf{w})$ might be difficult to optimize. Such a solving strategy has also been proven to own many good theoretical properties, like convergence and stability, under certain conditions.

\section{SPL Model and Algorithm Revisit}

\subsection{Axiomic Definition of SP-regularizer}

By mathematically abstracting the insightful properties underlying a
SPL regime, \cite{SPaR,SPMF} presented a formal definition for the
SP-ragularizer $f(v,\lambda )$ involved in the SPL model (\ref{eq:selfpaced_obj}) as follows:

\begin{definition}[SP-regularizer]\label{de:SP-regularizer}
	\label{SP_function_def}Suppose that $v$ is a weight variable, $\ell $ is the
	loss, and $\lambda $ is the age parameter. $f(v,\lambda )$ is
	called a self-paced regularizer, if
	
	\begin{enumerate}
		\item $f(v,\lambda )$ is convex with respect to $v\in \lbrack 0,1]$;		
		\item $v^{\ast }(\ell,\lambda)$ is monotonically decreasing with respect
		to $\ell $, and it holds that $\lim_{\ell \rightarrow 0}v^{\ast }(\ell,\lambda )=1$, $\lim_{\ell \rightarrow \infty }v^{\ast }(\ell,\lambda )=0$;		
		\item $v^{\ast }(\ell,\lambda)$ is monotonically increasing with respect
		to $\lambda $, and it holds that $\lim_{\lambda \rightarrow \infty }v^{\ast
		}(\ell,\lambda )\leq1$, $\lim_{\lambda \rightarrow 0}v^{\ast }(\ell,\lambda
		)=0$;
	\end{enumerate}
	where
	\begin{equation}\label{eq:SPAxiom}
	v^{\ast }(\ell,\lambda)=\arg \min_{v\in \lbrack 0,1]}v\ell
	+f(v,\lambda ).
	\end{equation}
\end{definition}

The three conditions in Definition \ref{SP_function_def} provide
basic principles for constructing a SP-regularizer. Condition 2 indicates that the model
inclines to select easy samples (with smaller losses) in favor of complex
samples (with larger losses). Condition 3 states that when the model
\textquotedblleft age\textquotedblright\ (controlled by the age parameter $%
\lambda$) gets larger, it tends to incorporate more, probably complex, samples to
train a \textquotedblleft mature\textquotedblright\ model. The
convexity in Condition 1 further ensures the soundness of this regularizer
for optimization.

Under this definition, multiple SP-regularizers have been
constructed. The following lists several typical ones, together with their
closed-form solutions $v^{\ast }(\lambda ,\ell )$ as defined in
Definition \ref{SP_function_def}:
\begin{equation}\label{eq:sp_regularizer}
	\begin{array}{l}
		f^{H}(v,\lambda )=-\lambda v\textrm{; }\ v^{\ast }(\ell,\lambda)=\left\{
		\begin{array}{c}
		1,\textrm{ if}\ \ell<\lambda  \\
		0,\textrm{ if}\ \ell\geq \lambda
		\end{array}%
		\right.
		\\
		f^{L}(v,\lambda )=\lambda (\frac{1}{2}v^{2}-v)\textrm{; }\
		v^{\ast }(\ell,\lambda) =\left\{
		\begin{array}{c}
		-\ell/\lambda +1,\textrm{ if}\ \ell<\lambda  \\
		0,\textrm{ if}\ \ell\geq \lambda
		\end{array}%
		\right.
		\\
		f^{M}(v,\lambda ,\gamma )=\frac{\gamma ^{2}}{v+\gamma
			/\lambda };\
		v^{\ast }(\ell, \lambda ,\gamma) =\left\{
		\begin{array}{c}
		1,\textrm{ if}\ \ell \leq \left( \frac{\lambda \gamma }{\lambda +\gamma }%
		\right) ^{2} \\
		0,\textrm{ if}\ \ell \geq \lambda ^{2} \\
		\gamma \left( \frac{1}{\sqrt{\ell }}-\frac{1}{\lambda }\right) ,\textrm{
			otherwise}.
		\end{array}%
		\right.
	\end{array}
\end{equation}

The above Eq. (\ref{eq:sp_regularizer}) represents the hard, linear and mixture SP-regularizers proposed in \cite{SPLVM}, \cite{SPaR}, and \cite{SPMF}, respectively. By using the AOS strategy to iteratively update $\mathbf{v}$ and $\mathbf{w}$ in the SPL regime (\ref{eq:selfpaced_obj}) with gradually increasing age parameter $\lambda$, a rational solution to the problem is expected to be progressively approached.

\subsection{Revisit AOS Algorithm for Solving SPL}

For convenience of notions, we briefly write $L(y_{i},g(\mathbf{x}_{i},%
\mathbf{w}))$ as $\ell_{i}(\mathbf{w})$/$\ell_{i}$ and $L(y,g(\mathbf{x},%
\mathbf{w}))$ as $\ell(\mathbf{w})$/$\ell$ in the following.

Given a SP-regularizer $f(v,\lambda)$, we can get the integrative function of $v^{\ast }(\ell,\lambda)$ calculated by Eq. (\ref{eq:SPAxiom}) as:
\begin{equation}
F_{\lambda }(\ell)=\int_{0}^{\ell}v^{\ast }(l,\lambda)dl. \label{eq:InteFun}
\end{equation}
The following result can then be proved. The proof is listed in the Appendix section.

\begin{theorem}\label{theorem1}
	For $v^{\ast }(\ell,\lambda)$ conducted by an SP-regularizer and $F_{\lambda }(\ell)$ calculated by (\ref{eq:InteFun}), given a fixed $\mathbf{w}^{\ast }$, it holds that:
\begin{equation*}
F_{\lambda }(\ell (\mathbf{w})) \leq Q_{\lambda }(\mathbf{w|w}^{\ast }) = F_{\lambda }(\ell (\mathbf{w}^{\ast }))+v^{\ast }(\ell (\mathbf{w%
}^{\ast }),\lambda)(\ell (\mathbf{w})\mathbf{-}\ell (\mathbf{w}^{\ast })).
\end{equation*}
\end{theorem}

The theorem can be easily understood by the fact that: Since $v^{\ast }(\ell,\lambda)$ is monotonically decreasing in $\ell$ based on Condition 2 of SP-regularizer Definition \ref{de:SP-regularizer}, its integrative $F_{\lambda }(\ell)$ is concave with respect to $\ell$, and thus it is easy to deduce that its Taylor series to $1^{st}$ order forms a upper bound of $F_{\lambda }(\ell)$.

Theorem \ref{theorem1} verifies that $Q_{\lambda }(\mathbf{w|w}^{\ast })$ represents a
tractable surrogate for $F_{\lambda }(\ell(\mathbf{w}))$. Specifically, only considering the terms with respect to $\mathbf{w}$, $Q_{\lambda }(\mathbf{w|w}^{\ast })$
simplifies $F_{\lambda }$, no matter how complicated its format is, as an easy weighted loss form
$v^{\ast }(\ell (\mathbf{w}^{\ast }),\lambda)\ell(\mathbf{w})$. This constitutes the fundament of our new understanding on the AOS algorithm for solving SPL.

Based on Theorem \ref{theorem1}, denote
\[
Q_{\lambda }^{(i)}(\mathbf{w|w}^{\ast })=F_{\lambda }(\ell_{i}(\mathbf{w}^{\ast
}))+v^{\ast }(\ell_{i}(\mathbf{w}^{\ast }),\lambda)(\ell_{i}(\mathbf{w})\mathbf{-}
\ell_{i}(\mathbf{w}^{\ast }),
\]
and we can then easily get that:
\begin{equation}\label{latentSPL}
\sum_{i=1}^{n}F_{\lambda }(\ell_{i}(\mathbf{w}))\leq \sum_{i=1}^{n}Q_{\lambda }^{(i)}(\mathbf{w|w}^{\ast }).
\end{equation}

Then we can prove the equivalence between the AOS strategy for solving the SPL
problem (\ref{eq:selfpaced_obj}) and the MM algorithm for solving $\sum_{i=1}^{n}F_{\lambda }(\ell_{i}(\mathbf{w%
}))$ under surrogate function $\sum_{i=1}^{n}Q_{\lambda }^{(i)}(\mathbf{w|w}%
^{\ast })$ as follows:

Denote $\mathbf{w}^{k}$ as the model parameters in the $k^{th}$ iteration of the
AOS implementation on solving SPL, and then its two alternative search steps in the next iteration can
be precisely explained as a standard MM scheme:

\emph{Majorization step}: To obtain each $Q_{\lambda }^{(i)}(\mathbf{w|w}^{k})$, we
only need to calculate $v^{\ast }(\ell_{i}(\mathbf{w}^{k}),\lambda)$
by solving the following problem under the corresponding SP-regularizer $ f(v_{i},\lambda )$:
$$v^{\ast }(\ell_{i}(\mathbf{w}^{k}),\lambda)=\underset{v_{i}\in \lbrack 0,1]}{\min }v_{i}\ell_{i}(\mathbf{w}^{k})+f(v_{i},\lambda ).$$

\noindent
This exactly complies with AOS step in updating $\mathbf{v}$ in (\ref{eq:selfpaced_obj}) under fixed $\mathbf{w}$.

\emph{Minimization step}: we need to calculate:%
\small
\begin{equation*}
	\begin{split}
	\mathbf{w}^{k+1}&=\arg \underset{\mathbf{w}}{\min }\sum_{i=1}^{n}F_{%
		\lambda }(\ell_{i}(\mathbf{w}^{k}))+v^{\ast }(\ell_{i}(\mathbf{w}%
	^{k}),\lambda)(\ell_{i}(\mathbf{w})\mathbf{-}\ell_{i}(\mathbf{w}^{k})) \\
	&=\arg\underset{\mathbf{w}}{\min }\sum_{i=1}^{n}v^{\ast }(\ell_{i}(%
	\mathbf{w}^{k}),\lambda)\ell_{i}(\mathbf{w}),
	\end{split}	
\end{equation*}
\normalsize
which is exactly equivalent to the AOS step in updating $\mathbf{w}$ in (\ref{eq:selfpaced_obj}) under fixed $\mathbf{v}$.

It is then easy to see that the commonly utilized AOS strategy in
previous SPL regimes is exactly the well known MM algorithm on a
minimization problem of the \emph{latent SPL objective} $\sum_{i=1}^{n}F_{\lambda }(\ell_{i}(\mathbf{w%
}))$ with the \emph{latent SPL loss} $F_{\lambda }(\ell(\mathbf{w}))$. Various off-the-shelf theoretical results of MM can then be readily employed
to explain the properties of such SPL solving strategy. For example, based on
the MM theory, the lower-bounded latent SPL objective is monotonically
decreasing during MM/AOS iteration, and the convergence of the SPL algorithm can then be guaranteed.

The above theory provides a new viewpoint for understanding SPL insight. More in-depth knowledge on SPL is then expected to be
extracted from it.

\subsection{Revisit SPL Model}

Now we try to discover more interesting insights from the latent SPL objective. To this aim, we first calculate the latent SPL losses under hard, linear and
mixture SP-regularizers, as introduced in (\ref{eq:sp_regularizer}), by Eq. (\ref{eq:InteFun}) as follows:
\begin{equation}
	\begin{array}{l}
		F_{\lambda }^{H}(\ell) =\left\{
		\begin{array}{cl}
			\ell, & \ell<\lambda , \\
			\lambda , & \ell\geq \lambda ;
		\end{array}
		\right.
		\\
		F_{\lambda }^{L}(\ell) =\left\{
		\begin{array}{cl}
		\ell-\ell^{2}/2{\lambda }, & \ell<\lambda , \\
		\lambda /2, & \ell\geq \lambda ;
		\end{array}
		\right.  \label{eq:Flambda}\ \ \ \ \ \ \ \ \ \ \ \ \\
		F_{\lambda ,\gamma }^{M}(\ell) =\left\{
		\begin{array}{cl}
		\ell, & \ell<\frac{1}{(1/\lambda +1/\gamma )^{2}}, \\
		\gamma (2\sqrt{\ell}-\ell/\lambda )-\frac{\gamma }{(1/\lambda +1/\gamma )}, &
		\frac{1}{(1/\lambda +1/\gamma )^{2}}\leq \ell<\lambda ^{2}, \\
		\gamma (\lambda -\frac{1}{1/\lambda +1/\gamma }), & \ell\geq \lambda ^{2}.
		\end{array}
		\right.
	\end{array}
\end{equation}

\begin{figure*}
\scalebox{0.41}[0.41]{\includegraphics{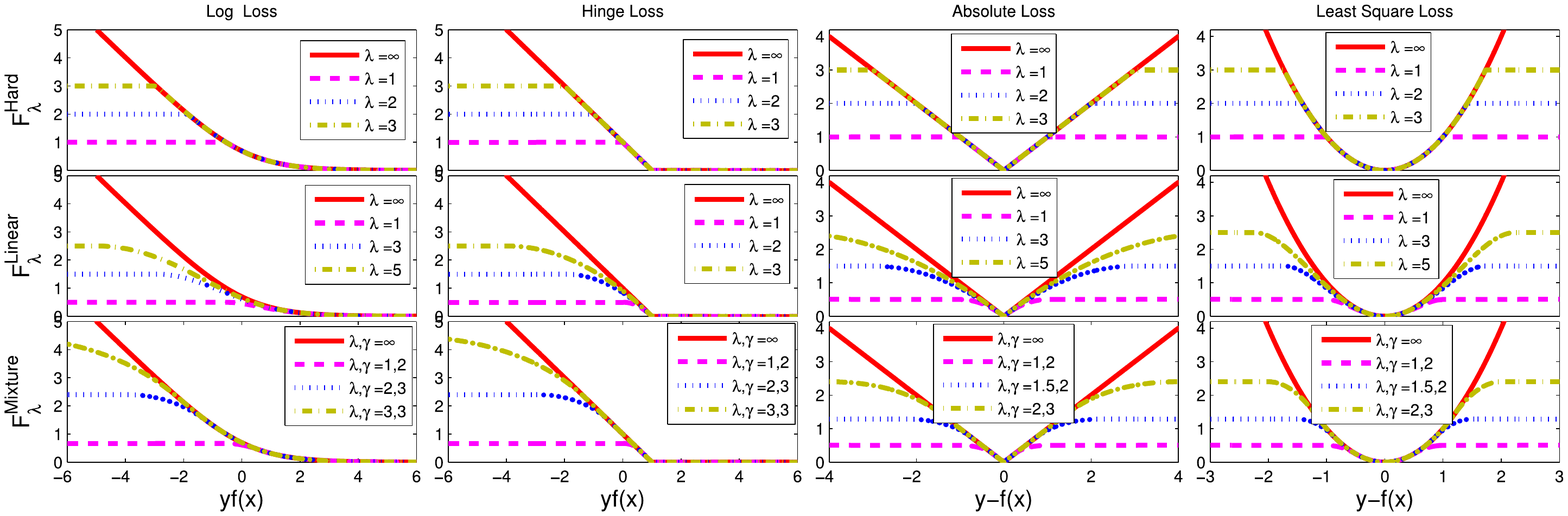}}
	\caption{Graphical illustration for latent SPL losses $F_{\lambda }^{Hard}(\ell)$,  $F_{\lambda }^{Linear}(\ell)$ and $F_{\lambda,\gamma }^{Mixture}(\ell)$ conducted by the hard, soft and mixture SP-regularizers on different loss functions, including the logistic loss, the hinge loss, the absolute loss and the least square loss, under various pace parameters in 1-dimensional cases, respectively. Note that when $\lambda=\infty$ ($\lambda,\gamma=\infty$ in mixture cases), the latent SPL loss $F_{\lambda }(\ell)$ degenerates to the original loss $\ell$.
	}\label{fig:F}
\end{figure*}

The configurations of these $F_{\lambda }(\ell)$s under different age parameters are depicted in Figure \ref{fig:F} for easy observation.

Some common patterns under these latent SPL losses can be easily observed from
Figure \ref{fig:F}. E.g., there is an evident suppressing effect of $F_{\lambda
}(\ell)$ on large losses as compared with the original loss function $\ell$. When $\ell$ is larger
than a certain threshold, $F_{\lambda }(\ell)$ will become a constant
thereafter. This provides a rational explanation on why the SPL regime can perform robust in the
presence of extreme outliers or heavy noises: The samples with loss
values larger than the age threshold will have little influence to the model
training due to their 0 gradients. Corresponding to the
original SPL model, these large-loss samples will be with 0 importance weights $v_{i}$,
and thus have no effect on the optimization of model parameters.

Now, let's reexamine the intrinsic mechanism inside SPL
implementation based on such understanding. In the beginning of SPL iteration, the age $\lambda $ is small, the latent loss
function $F_{\lambda }(\ell)$ has a significant suppressing effect on large
losses and only allows small amount of high-confidence samples (with small
loss values) into training; then with gradually increasing $\lambda$, the
suppressing effect of $F_{\lambda }(\ell)$ will gradually become
weaker and more relatively less informative samples incline to be involved
into training. Through such robust guidance, more and more faithful
data knowledge tend to be incrementally learned by such learning scheme. Such a gradually
changing tendency of latent SPL loss $F_{\lambda }(\ell)$ can be easily
understood by seeing Figure \ref{fig:F}.

\subsection{Relationship with NCRP}

An interesting observation is that the latent SPL objective $F_{\lambda }(\ell)$
has a close relationship to NCRP widely investigated in machine learning and
statistics. E.g., the hard and linear SPL objectives $F_{\lambda
}^{H}(\ell)$ and $F_{\lambda }^{L}(\ell)$ comply exactly with the forms of CNP and MCP, as defined in Eq. (\ref{eq:NCRP}), imposed on $\ell$ by setting $\gamma =1$, respectively. I.e.,
\begin{equation*}
F_{\lambda }^{H}(\ell)=p_{1,\lambda }^{CNP}(\ell),\ \ F_{\lambda }^{L}(\ell)=p_{1,\lambda
}^{MCP}(\ell).
\end{equation*}

\noindent
Furthermore, the form of $F_{\lambda ,\gamma }^{M}(\ell)$ is almost similar to the
SCAD term, both containing three phases of values, and the first and third of both are linear
and constant, respectively. The only difference is in the second phase, where $F_{\lambda ,\gamma }^{M}(\ell)$ is of linear+sqrt+constant form while SCAD is of a linear+square+constant one. Actually, it is easy to deduce that any $F_{\lambda
}(\ell)$ led by a SP-ragularizer is non-convex, and has a very similar
configuration with a general NCRP. Such a natural relationship on one hand
provides a new viewpoint to see NCRP and facilitates more choices of NCRP formulations by virtue of $F_{\lambda }(\ell)$ obtained under various SP-regularizers, and on the other hand inspires us to borrow mature
statistical tools and theoretical results on NCRP to further understand SPL
insight in our future investigation.

This relationship is also helpful for finding self-paced formats of more typical NCPR terms. Here we also deduce the SP-regularizers of another two commonly utilized NCPR terms \cite{ZhangNonconvex}: LOG and EXP (see Eq. (\ref{eq:NCRP})).

For LOG, we can construct the following SP-regularier:%
\begin{eqnarray*}
f^{LOG}(v,\lambda ,\alpha ) =\frac{1}{\alpha }KL(1+\alpha \lambda ,v)
=\frac{1}{\alpha }\left((1+\alpha \lambda )\log \frac{1+\alpha \lambda }{v}-%
(1+\alpha \lambda)+v\right),
\end{eqnarray*}%
where $KL(x ,y)$ denotes the Kullback-Leibler (KL) distance~\cite{ZhangNonconvex} between two variables. As calculated by Eq. (\ref{eq:SPAxiom}),
its optimal weight $v^{\ast }(\ell, \lambda ,\alpha)$ is:
\begin{equation*}
v^{\ast }(\ell, \lambda ,\alpha)=\left\{
\begin{array}{cc}
1 & \ell \leq \lambda,  \\
\frac{1+\alpha \lambda }{1+\alpha \ell } & \ell >\lambda .%
\end{array}%
\right.
\end{equation*}%
It is easy to prove that such defined LOG SP-regularizer complies with the
three conditions in Definition \ref{de:SP-regularizer}. By virtue of Eq. (\ref{eq:InteFun}), we can obtain its latent SPL
loss with the form:%
\begin{eqnarray}
F^{LOG}(\ell, \lambda ,\alpha ) =\left\{
\begin{array}{cc}
\ell  & \ell \leq \lambda,  \\
p^{LOG}_{1 +\alpha \lambda ,\alpha}(\ell)+C_{\lambda ,\alpha } &
\ell >\lambda ,%
\end{array}%
\right. \label{LOGLoss}
\end{eqnarray}

\noindent
where $C_{\lambda ,\alpha }=\lambda -\frac{1+\alpha \lambda }{\alpha }\log
(1+\alpha \lambda )$ is a constant independent of $\ell$ and $p^{LOG}_{\gamma,\alpha}(\cdot)$ is defined as Eq. (\ref{eq:NCRP}).

Besides, the EXP SP-regularizer can be constructed as::%
\begin{eqnarray*}
f^{EXP}(v;\lambda ,\alpha ) =\frac{1}{\alpha }KL(v,\exp (\alpha \lambda ))
=\frac{1}{\alpha }(v\log \frac{v}{\exp (\alpha \lambda )}-v+\exp (\alpha
\lambda )).
\end{eqnarray*}%

\noindent
Its optimal importance weight can be calculated by Eq. (\ref{eq:SPAxiom}) as:%
\begin{equation*}
v^{\ast }(\ell,\lambda ,\alpha )=\left\{
\begin{array}{cc}
1 & \ell \leq \lambda,  \\
\exp (-\alpha (\ell -\lambda )) & \ell >\lambda .%
\end{array}%
\right.
\end{equation*}%

\noindent
The corresponding latent SPL loss is:%
\begin{equation}
F^{EXP}(\ell ,\lambda ,\alpha ) =
\left\{
\begin{array}{cc}
\ell  & \ell \leq \lambda,  \\
p^{EXP}_{\alpha,\exp (-\alpha \lambda)}(\ell)+C_{\lambda
,\alpha } & \ell >\lambda ,%
\end{array}%
\right.\label{EXPLoss}
\end{equation}%

\noindent
where $C_{\lambda ,\alpha }=\lambda +\frac{1}{\alpha }-\frac{\exp (\alpha
\lambda )}{\alpha }$ and $p^{EXP}_{\gamma,\alpha}(\cdot)$ is defined as Eq. (\ref{eq:NCRP}).

It should be noted that LOG and EXP are different from CNP, MCP and SCAD in their large-loss-suppressing effects. The latter ones suppress large losses as a constant (see Eq. (\ref{eq:NCRP})) while the former do this task by the gradually more slowly increasing property of logarithmic and negative exponential functions.
It is easy to see that in large loss cases, the LOG and EXP latent SPL loss functions (\ref{LOGLoss}) and (\ref{EXPLoss}) degenerate to the conventional LOG and EXP terms, and thus possess similar robust mechanism on suppressing outliers/heavy noises. Through properly adjusting age parameter $\lambda$, they are capable of adapting different noise extents in data.

\subsection{Loss-prior-embedding Property of SPL}

An intrinsic property of SPL is to
decompose the minimization of the robust but difficult-to-solve non-convex
loss $F_{\lambda }(\ell(\mathbf{w}))$ into two easier optimization
problems with \mbox{respect} to sample importance weights $\mathbf{v}$\ (solved by
the closed-form solution to SP-regularizer) and model parameters\ $\mathbf{w}
$\ (solved by weighted loss problem). Such decomposition not only simplifies
the solving of the problem as an easy re-weighted strategy~\cite{Reweight}, but makes
it feasible to embed helpful loss prior knowledge into an SPL scheme. Specifically, since the sample importance weight imposed on a sample in SPL model reflects the extent of its loss value based on Condition 2 of Definition \ref{de:SP-regularizer} for SP-regularizer (the larger the loss, the smaller the weight), the loss priors can be readily encoded as a regularization term or a constraint on $\mathbf{v}$ to deliver such knowledge in the SPL model.

In practical cases, there always exist some loss priors which can be easily obtained from training data before the learning process. Here we list some typical ones as follows:

\begin{enumerate}
	\item Outlier prior: Some samples are significantly deviated from the main part of data sets, and thus they should be with extremely large losses.

	\item Spatial/temporal smoothness prior: Some spatially/temporally adjacent
	samples tend to be with relatively similar large/small losses.
	
	\item Sample importance order prior: a sample is preknown to be with smaller loss value
	 (i.e., cleaner, easier, more high-confident) than the other one.
	
	\item Diversity prior: Meaningful samples, which should be learned with small loss values (i.e., capable of being predicted accurately) for the learning task, should be scattered across the data range so that the learning can possibly include
	global-scale data knowledge.
\end{enumerate}

All these loss prior knowledge can be embedded into an SPL scheme by properly encoding $\mathbf{v}$. E.g., Prior 1 can be realized by directly constraining the importance weights $v_i$ of those outliers to be zeroes; Prior 2 can be formulated as a graph
Laplacian term $\mathbf{v}^{T}\mathbf{Lv}$, where $\mathbf{L}$\ is the
Laplacian matrix on the data adjacent matrix; Prior 3 can be
encoded as supplemental constraint $v_{i}>v_{j}$ if the $i^{th}$ sample is known more
cleaner/easier than $j^{th}$ one \cite{SPCL}; and Prior 4 can be realized by a $-l_{2,1}$ norm or $-l_{0.5,1}$
norm (anti-group-sparsity) on $\mathbf{v}$, as utilized in \cite{SPMIL} and \cite{SPLD}, respectively.

Let's see this loss-prior-embedding property from the perspective of NCPR. Currently various elegant solving strategies have been designed for solving a general or specific NCPR problem~\cite{ZhangTongNonconvex} so as to approach a local minimum or stationary point of the problem. In most of these strategies, however, it has been neglected whether the obtained solution complies with some \mbox{evident} loss prior knowledge. E.g., by using certain techniques, we might obtain a local minimum of the investigated non-convex problem. However, it might occur that the loss of Sample A predicted at this local minimum is larger than that of Sample B, while we have an intuitive or easily-obtained prior that A is much noisy than B. This implies that this solution, albeit being a local minimum, is an irrational one to the problem. If we transform this NCPR problem into a SPL regime based on the analysis provided in Section 3.4, and readily encode such loss priors (e.g., sample importance order loss prior) into the latent variables $\mathbf{v}$ as a regularizer or a constraint to the problem, the obtained solution is expected to be more easily avoiding such unreasonable local minima violating the apriori loss priors. Such easy loss-prior embedding capability thus inclines to guide a sounder learning manner for NCPR as well as SPL, which also might provide a new viewpoint of alleviating the local-minimum-issue existed in NCPR problems.

\subsection{SPCL Revisit}
Self-paced curriculum learning (SPCL)~\cite{SPCL} was proposed to relate CL and SPL from the viewpoint of human learning. As opposed to ``instructor-driven" and ``student-driven" learning manners as CL and SPL, respectively, through involving prior knowledge of sample importance into SPL iteration progress, SPCL is analogous to a more rational ``instructor-student-collaborative" learning mode like practical human education.

SPCL actually illuminates the fundament of the SPL regime embedded with loss priors in the perspective of cognitive science. Specifically, the curriculum knowledge in SPCL complies with the loss priors in this study. That is, a teacher might know some curriculum information to guide the learning process of a student, e.g., some curriculum is meaningless to learn (outlier prior); multiple curriculums are closely related and should be learned jointly (smoothness prior); one curriculum is much more difficult than another and thus should be learned first (sample importance order prior); diverse curriculums should be learned together to make the knowledge possibly comprehensive (diversity prior). Such relationship on one hand facilitates a natural interpretation for the empirical effectiveness of SPCL by the fact that the embedded loss priors (curriculums) helps alleviate the local-minimum-issue of the underlying non-convex optimization problems and guarantee a sound robust learning, and on the other hand illustrates that a SPL scheme with properly specified loss priors more accords to a rational human education manner in real life as compared to the pure CL or SPL strategies.

\section{FCVID Experiment}

For a real large-scale pattern recognition task, with limited human labor and computation resources, in general we can only get weakly labeled samples, containing large amount of low-confidence annotations (with wrong or uncertain labels). This often leads to a noisy training dataset, which inclines to hamper the robustness of the utilized learning algorithm. The SPL strategy is thus appropriate to be employed to alleviate this issue. Highly deviated samples (i.e., with relatively larger loss values) will be automatically screened out (i.e., with zero-valued $v_i$) from training and not negatively influence the learning quality, while those samples with high-confidence annotations (i.e., with smaller loss values) tend to be selected and gradually rectify the learning performance. However, as we have analyzed, SPL intrinsically corresponds to solving a non-convex optimization problem, and useful loss priors are thus required to help avoid the problem stuck to irrational local minima. Through ameliorating the sample importance loss prior, we introduce an advanced group-partial-order loss prior, which is especially useful in such large-scale weakly labeled scenarios. Specifically, we can construct this loss prior in two steps: first group and rank data based on the difficulty for annotating them, and then impose a hierarchy loss structure by letting samples located in groups in front of the ranking list (i.e., with higher-confidence labels) with a larger weights than those in the behind. The SPL regime is then expected to be soundly guided under such loss priors.

For verification, we use a real-world big dataset called Fudan-columbia Video Dataset (FCVID)~\cite{jiang2015exploiting}, which is by far one of the biggest annotated video set~\cite{jiang2015exploiting} and thus is challenging for conventional concept detection techniques.
Our goal is to learn detectors that can automatically recognize concepts occurring in the video content, such as people, objects, actions, etc. FCVID contains $91,223$ YouTube videos ($4,232$ hours) from $239$ categories. The class covers a wide range of concepts like activities, objects, scenes, sports, DIY, etc. Each video is manually labeled to one or more categories. As manually labeled videos are difficult to collect, we train concept detectors only using the contexual information about video such as their titles, descriptions and latent topics. For each concept, a video is automatically labeled as a positive sample if the concept name can be found in its video metedata. The generated weak labels are noisy and have both low accuracy and low recall: the labeled concepts may not present in the video content whereas concepts not in the web label may well appear in the video. The ground truth labels are only used in test to evaluate the performance. The performance is evaluated in terms of the precision of the top 5 and 10 ranked videos (P@5 and P@10) and mean Average Precision (mAP) of 239 concepts.

We extract the Convolutional Neural Network (CNN), specifically AlexNet~\cite{Alexnet}, features over each keyframe and create video-level feature by average pooling. The features are used across all methods. We build our method on top of the CNN features and the $l_2$-regularized hinge loss is used as our loss function. The AOS algorithm is used to solve the optimization problem. To construct the group-partial-order loss prior knowledge into SPL on this task, we cluster the videos into a number of latent topics based on their metadata. Then we rank these groups in the ascending order of the distance between cluster center to the entire concept class center. Samples located in clusters in front of the ranking list should correspond to more high-confident ones and incline to have a larger weights $v_i$ (i.e., with smaller loss values) than those ranking backwards. The hard SP-regularizer~\cite{SPLVM} was used in the SPL scheme. In implementation, we used 10\% of top-ranked samples in the first iteration to get initialization and stopped increasing the model age $\lambda$ after 100 about iterations.

We compare our method against the following baseline methods, which cover both the classical and the state-of-the-art algorithms on the same problem. Besides, the comparison between baseline helps us understand the contribution of the loss prior knowledge on this problem. \textit{BatchTrain} trains a single SVM model using all videos with noisy labels. \textit{AdaBoost} is a classical ensemble approach that combines the sequentially trained base classifiers in a weighted fashion~\cite{friedman2002stochastic}. \textit{Self-Paced Learning (SPL)} is the original SPL method without considering loss prior knowledge~\cite{SPLVM}. \textit{BabyLearning} is a recently proposed method that simulates baby learning by starting with few training samples and fine-tuning using more weakly labeled videos crawled from the search engine~\cite{liang2015towards}. \textit{GoogleHNM} is a hard negative mining method proposed by Google~\cite{varadarajan2015efficient}. It utilizes hard negative mining to train the mixture of experts model according to the video's YouTube topics. The hyper-parameters of all methods including the baseline methods are tuned on the same standard validation set.

Table~\ref{tab:exp_overview} compares the precision and mAP of different
methods. As we see, the SPL method with group-partial-order loss priors achieves
the state-of-the-art result which outperforms the recently proposed
methods BabyLearning~\cite{liang2015towards} and
GoogleHNM~\cite{varadarajan2015efficient}. The average improvement over
baseline method on 239 classes are statistically significant at $p$-level
of $0.05$. As compared to classical methods such as BatchTrain and Adaboost,
the results empirically demonstrate the benefit of the robustness mechanism underlying SPL.
Besides, our improvement over the standard SPL method suggests that
incorporating loss prior knowledge into learning yield a significant boost.

\begin{table}[ht]
\centering
\caption{Performance comparison of the proposed and baseline methods on FCVID.}
\label{tab:exp_overview}
\vspace{3mm}
\setlength{\tabcolsep}{4pt} 
\renewcommand{\arraystretch}{1.1} 
\begin{tabular}{|l||c|c|c|c|}
\hline
Method      & P@5 & P@10  & mAP\\ \hline \hline
BatchTrain             & 0.782 & 0.763 & 0.469 \\
Adaboost~\cite{friedman2002stochastic}              & 0.211 & 0.173 &
0.08 \\
SPL~\cite{SPLVM}           &  0.793 & 0.754  & 0.414 \\
GoogleHNM~\cite{varadarajan2015efficient}      &  0.781 &  0.757 & 0.472
\\
BabyLearning~\cite{liang2015towards}   & 0.834  & 0.817  & 0.496 \\
\textbf{SPL with partial-order-loss-prior} &    \textbf{0.889}    &   \textbf{0.874}   &
\textbf{0.5329}\\\hline
\end{tabular}
\end{table}
\section{Conclusion}

We have provided some new insightful understanding to the conventional SPL regime in this study. On one hand, we have shown that the AOS algorithm generally utilized for solving SPL exactly complies with the known MM algorithm on a latent SPL objective, and on the other hand we have verified that the loss function contained in this latent SPL objective precisely accords with the famous non-convex regularized penalty (NCRP). The effectiveness, especially its robustness to outliers/heavy noises, of SPL, as substantiated by previous experiences, can then be naturally explained under such understanding. We also analyzed the superiority of SPL on its easy loss-prior-embedding property, which provides a new methodology for alleviating the local-minimum-issue in general NCRP optimization problems. In our future investigation, we will attempt to employ the theories on MM and NCRP to more deeply explore the theoretical/statistical properties underlying the SPL regimes.

\section*{Appendix: Proof of Theorem 1}
To prove the theorem, we need to show that
$$F_{\lambda}(\ell)\leq F_{\lambda}(\ell_0)+v^{\ast}(\ell_0,\lambda)(\ell-\ell_0).$$
There are two cases should be dealt with.

1. $v^{\ast}(\ell,\lambda)$ is continous with respect to $\ell$.

From Eq. (3), we have that $$v^{\ast}(\ell,\lambda)=F_{\lambda}^{\prime}(\ell).$$
By Definition 1, $v^{\ast}(\ell,\lambda)\geq0$ when $\ell\geq0$, and thus $F_{\lambda}^{\prime}(\ell)$ is nondecreasing with respect to $\ell$ on $[0,\infty)$. Besides, $v^{\ast}(\ell,\lambda)$ is monotonically decreasing with respect to $\ell$. Therefore, we can conclude that $F_{\lambda}(\ell)$ is concave on $[0,\infty)$. Based on the property of concave function, we have
$$F_{\lambda}(\ell)\leq F_{\lambda}(\ell_0)+F_{\lambda}^{\prime}(\ell_0)(\ell-\ell_0)=F_{\lambda}(\ell_0)+v^{\ast}(\ell_0,\lambda)(\ell-\ell_0).$$

2. $v^{\ast}(\ell,\lambda)$ is discontinuous with respect to $\ell$.

With out loss of generality, suppose there is only one discontinuous $\tilde{\ell}\in[0,\infty)$. When $\ell,\ell_0\in[0,\tilde{\ell})$ or $\ell,\ell_0\in(\tilde{\ell},\infty)$, following the similar derivation, we also have that
$$F_{\lambda}(\ell)\leq F_{\lambda}(\ell_0)+v^{\ast}(\ell_0,\lambda)(\ell-\ell_0)$$
holds.

Now suppose $\ell\in[0,\tilde{\ell})$ and $\ell_0\in(\tilde{\ell},\infty)$. Pick $\ell_1\in[0,\tilde{\ell})$, and then we have that
$$F_{\lambda}(\ell)\leq F_{\lambda}(\ell_1)+v^{\ast}(\ell_1,\lambda)(\ell-\ell_1),$$
and
$$F_{\lambda}(\tilde{\ell})\leq F_{\lambda}(\ell_0)+v^{\ast}(\ell_0,\lambda)(\tilde{\ell}-\ell_0).$$
Denote $v^{\ast}(\tilde{\ell},\lambda)^{-}=\lim_{\ell\rightarrow \tilde{\ell}^{-}}v^{\ast}(\ell,\lambda)$, and let $\ell_1\rightarrow\tilde{\ell}^{-}$. Since $F_{\lambda}(\ell)$ is continuous, we can have that
$$F_{\lambda}(\ell)\leq F_{\lambda}(\tilde{\ell})+v^{\ast}(\tilde{\ell},\lambda)^{-}(\ell-\tilde{\ell}).$$
Therefore,
\begin{equation*}
	\begin{split}
		F_{\lambda}(\ell)-F_{\lambda}(\ell_0)&=F_{\lambda}(\ell)-F_{\lambda}(\tilde{\ell})+F_{\lambda}(\tilde{\ell})-F_{\lambda}(\ell_0)\\
		&\leq v^{\ast}(\lambda;\tilde{\ell})^{-}(\ell-\tilde{\ell})+v^{\ast}(\ell_0,\lambda)(\tilde{\ell}-\ell_0)\\
		&\leq v^{\ast}(\ell_0,\lambda)(\ell-\tilde{\ell})+v^{\ast}(\ell_0,\lambda)(\tilde{\ell}-\ell_0)\\
		&=v^{\ast}(\ell_0,\lambda)(\ell-\ell_0),
	\end{split}	
\end{equation*}
where the second inequality holds due to the fact that $\ell\leq\tilde{\ell}$ and $v^{\ast}(\ell,\lambda)\geq0$ is decreasing with respect to $\ell$.

Similarly, if $\ell\in(\tilde{\ell},\infty)$ and $\ell_0\in[0,\tilde{\ell})$, the result also hods.

Now we consider the case $\ell_0=\tilde{\ell}$. Suppose $\ell\in[0,\tilde{\ell})$ (derivation is similar for $\ell\in(\tilde{\ell},\infty)$), and pick $\ell_1\in[0,\tilde{\ell})$. We have that
$$F_{\lambda}(\ell)\leq F_{\lambda}(\ell_1)+v^{\ast}(\ell_1,\lambda)(\ell-\ell_1).$$
Let $\ell_1\rightarrow\tilde{\ell}^{-}$. Since $F_{\lambda}(\ell)$ is continuous, we can have that
$$F_{\lambda}(\ell)\leq F_{\lambda}(\ell_0)+v^{\ast}(\lambda;\tilde{\ell})^{-}(\ell-\ell_0)\leq F_{\lambda}(\ell_0)+v^{\ast}(\ell_0,\lambda)(\ell-\ell_0),$$
where the second inequality holds due to the fact that $\ell\leq \ell_0$ and $v^{\ast}(\ell,\lambda)\geq0$ is decreasing with respect to $\ell$.

From the above discussion, we can conclude that $$F_{\lambda}(\ell)\leq F_{\lambda}(\ell_0)+v^{\ast}(\ell_0,\lambda)(\ell-\ell_0).$$ Substitute $\ell$ and $\ell_0$ with $\ell({\bf w})$ and $\ell({\bf w}^{\ast})$, respectively, and then Theorem 1 follows.


\section*{References}

\begin{thebibliography}{10}

\bibitem{CLA2}
S.~Basu and J.~Christensen.
\newblock Teaching classification boundaries to humans.
\newblock In {\em AAAI}, 2013.

\bibitem{CL}
Y.~Bengio, J.~Louradour, R.~Collobert, and J.~Weston.
\newblock Curriculum learning.
\newblock In {\em ICML}, 2009.

\bibitem{EXP}
P.~S. Bradley and O.~L. Mangasarian.
\newblock Feature selection via concave minimization and support vector
  machines.
\newblock In {\em ICML}, 1998.

\bibitem{Reweight}
E.~J. Cand\`{e}s, M.~B. Wakin, and S.P. Boyd.
\newblock Enhancing sparsity by reweighted l1 minimization.
\newblock {\em Journal of Fourier Analy-sis and Applications},
  14(5-6):877--905, 2008.

\bibitem{SCAD}
J.~Fan and R.~Li.
\newblock Variable selection via nonconcave penalized likelihood and its oracle
  properties.
\newblock {\em Journal of American Statistical Association},
  96(456):1348--1360, 2001.

\bibitem{friedman2002stochastic}
J.~H Friedman.
\newblock Stochastic gradient boosting.
\newblock {\em Computational Statistics \& Data Analysis}, 38(4):367--378,
  2002.

\bibitem{CAP3}
P.~Gong, C.~Zhang, Z.~Lu, J.~Huang, and J.~Ye.
\newblock A general iterative shrinkage and thresholding algorithm for
  non-convex regularized optimization problems.
\newblock In {\em ICML}, 2013.

\bibitem{SPaR}
L.~Jiang, D.~Meng, T.~Mitamura, and A.~Hauptmann.
\newblock Easy samples first: self-paced reranking for zeroexample multimedia
  search.
\newblock In {\em ACM MM}, 2014.

\bibitem{SPLD}
L.~Jiang, D.~Y. Meng, S.~Yu, Z.~Z. Lan, , S.~G. Shan, and A.~Hauptman.
\newblock Self-paced learning with diversity.
\newblock In {\em NIPS}, 2014.

\bibitem{SPCL}
L.~Jiang, D.~Y. Meng, Q.~Zhao, S.~G. Shan, and A.~Hauptman.
\newblock Self-paced curriculum learning.
\newblock In {\em AAAI}, 2015.

\bibitem{jiang2015exploiting}
Y.~G. Jiang, Z.~Wu, J.~Wang, X.~Xue, and S.~F. Chang.
\newblock Exploiting feature and class relationships in video categorization
  with regularized deep neural networks.
\newblock {\em arXiv preprint arXiv:1502.07209}, 2015.

\bibitem{NCPList}
Y.~Kang, Z.~Zhang, and W.~Li.
\newblock On the global convergence of majorization minimization algorithms for
  nonconvex optimization problems.
\newblock In {\em arXiv: 1504.07791v2}, 2015.

\bibitem{Khan}
F.~Khan, X.~Zhu, and B.~Mutlu.
\newblock How do humans teach: on curriculum learning and teaching dimension.
\newblock In {\em NIPS}, 2011.

\bibitem{Alexnet}
Alex Krizhevsky, Ilya Sutskever, and Geoffrey~E. Hinton.
\newblock Imagenet classification with deep convolutional neural networks.
\newblock In {\em NIPS}, 2012.

\bibitem{SPLVM}
M.~Kumar, B.~Packer, and D.~Koller.
\newblock Self-paced learning for latent variable models.
\newblock In {\em NIPS}, 2010.

\bibitem{SPSegmentation}
M.~Kumar, H.~Turki, D.~Preston, and D.~Koller.
\newblock Learning specific-class segmentation from diverse data.
\newblock In {\em ICCV}, 2011.

\bibitem{MM1}
K.~Lange, D.~Hunter, and I.~Yang.
\newblock Optimization transfer using surrogate objective functions.
\newblock {\em Journal of Computational and Graphical Statistics}, 9(1):1--20,
  2000.

\bibitem{CLAPP3}
A.~Lapedriza, H.~Pirsiavash, Z.~Bylinskii, and A.~Torralba.
\newblock Are all training examples equally valuable?
\newblock In {\em CoRR abs/1311.6510}, 2013.

\bibitem{SPVCD}
Y.~Lee and K.~Grauman.
\newblock Learning the easy things first: Self-paced visual category discovery.
\newblock In {\em CVPR}, 2011.

\bibitem{liang2015towards}
X.~Liang, S.~Liu, Y.~Wei, L.~Liu, and S.~Lin, L.and~Yan.
\newblock Towards computational baby learning: A weakly-supervised approach for
  object detection.
\newblock In {\em ICCV}, 2015.

\bibitem{CLApp1_}
E.~Ni and C.~Ling.
\newblock Supervised learning with minimal effort.
\newblock In {\em Advances in Knowledge Discovery and Data Mining}, pages
  476--487. Springer, 2010.

\bibitem{PR1}
Z.~Pan and C.~Zhang.
\newblock Relaxed sparse eigenvalue conditions for sparse estimation via
  non-convex regularized regression.
\newblock {\em Pattern Recognition}, 48(1):231¨C243, 2015.

\bibitem{CLApp1}
V.~I. Spitkovsky, H.~Alshawi, and D.~Jurafsky.
\newblock Baby steps: How ``less is more" in unsupervised dependency parsing.
\newblock In {\em NIPS}, 2009.

\bibitem{SPTrack}
J.~Supan\v{c}i\v{c}~III and D.~Ramanan.
\newblock Self-paced learning for long-term tracking.
\newblock In {\em CVPR}, 2013.

\bibitem{CAPLoss}
S.~Suzumura, K.~Ogawa, M.~Sugiyama, and I.~Takeuchi.
\newblock Outlier path: a homotopy algorithm for robust svm.
\newblock In {\em ICML}, 2014.

\bibitem{SPShift}
K.~Tang, V.~Ramanathan, F.~Li, and D.~Koller.
\newblock Shifting weights: Adapting object detectors from image to video.
\newblock In {\em NIPS}, 2012.

\bibitem{MM}
F.~Vaida.
\newblock Parameter convergence for \mbox{EM and MM} algorithms.
\newblock {\em Statistica Sinica}, 15(3):831, 2005.

\bibitem{varadarajan2015efficient}
B.n Varadarajan, G.~Toderici, S.~Vijayanarasimhan, and A.~Natsev.
\newblock Efficient large scale video classification.
\newblock {\em arXiv preprint arXiv:1505.06250}, 2015.

\bibitem{MCPLoss}
S.~Wang, D.~Liu, and Z.~Zhang.
\newblock Nonconvex relaxation approaches to robust matrix recovery.
\newblock In {\em IJCAI}, 2013.

\bibitem{ZhangTongNonconvex}
Z.~Wang, H.~Liu, and T.~Zhang.
\newblock Optimal computational and statistical rates of convergence for sparse
  nonconvex learning problems.
\newblock {\em Annals of Statistics}, 42:2164--2201, 2014.

\bibitem{LOG2}
J.~Weston, A.~Elisseeff, B.~Sch\"{o}lkopf, and M.~Tipping.
\newblock Use of the zero-norm with linear models and kernel methods.
\newblock {\em Journal of Machine Learning Research}, 3:1439--1461, 2003.

\bibitem{MED14}
S.~Yu, L.~Jiang, Z.~Mao, and et~al.
\newblock \mbox{CMU-I}nformedia@\mbox{TRECVID} 2014 multimedia eventdetection
  \mbox{(MED)}.
\newblock In {\em TRECVID Video Retrieval Evaluation Workshop}, 2014.

\bibitem{MCP}
C.~Zhang.
\newblock Nearly unbiased variable selection under minimax concave penalty.
\newblock {\em Annals of Statistics}, pages 894--942, 2010.

\bibitem{CAP2}
C.~Zhang and T.~Zhang.
\newblock A general theory of concave regularization for high-dimensional
  sparse estimation problems.
\newblock {\em Statistical Science}, 27(4):576--593, 2012.

\bibitem{SPMIL}
D.~Zhang, D.~Meng, and J.~Han.
\newblock Co-saliency detection via a self-paced multiple-instance learning
  framework.
\newblock In {\em ICCV}, 2015.

\bibitem{SPMIL-PAMI}
D.~Zhang, D.~Meng, and J.~Han.
\newblock Sp-mil: A self-paced multiple-instance learning framework for
  co-saliency detection.
\newblock {\em IEEE Transactions on Pattern Analysis and Machine Intelligence},
  2016.

\bibitem{CAP1}
T.~Zhang.
\newblock Analysis of multi-stage convex relaxation for sparse regularization.
\newblock {\em The Journal of Machine Learning Research}, 11:1081--1107, 2010.

\bibitem{ZhangNonconvex}
Z.~Zhang and B.~Tu.
\newblock Nonconvex penalization using laplace exponents and concave
  conjugates.
\newblock In {\em NIPS}, 2012.

\bibitem{SPMF}
Q.~Zhao, D.~Y. Meng, L.~Jiang, Q.~Xie, Z.~B. Xu, and A.~Hauptman.
\newblock Self-paced learning for matrix factorization.
\newblock In {\em AAAI}, 2015.

\end{thebibliography}
\end{document}